\documentclass{article}
\usepackage[table]{xcolor}
\usepackage{iclr2025_conference,times}

\usepackage{amsmath,amsfonts,bm}

\def\eqref#1{equation~\ref{#1}}
\def\Eqref#1{Equation~\ref{#1}}
\def\1{\bm{1}}

\def\vtheta{{\bm{\theta}}}

\DeclareMathAlphabet{\mathsfit}{\encodingdefault}{\sfdefault}{m}{sl}
\SetMathAlphabet{\mathsfit}{bold}{\encodingdefault}{\sfdefault}{bx}{n}

\def\gN{{\mathcal{N}}}

\newcommand{\E}{\mathbb{E}}

\newcommand{\R}{\mathbb{R}}

\usepackage{hyperref}
\usepackage{url}

\usepackage{booktabs}
\usepackage{amsfonts}
\usepackage{nicefrac}
\usepackage{microtype}
\usepackage{natbib}
\usepackage{hyperref}
\usepackage{enumitem}
\usepackage{float}
\usepackage{amsmath}
\usepackage{algorithm}
\usepackage{algpseudocode}
\usepackage{cleveref}
\usepackage{autonum}
\usepackage{graphicx}
\usepackage{multirow}
\usepackage{multicol}
\usepackage{dsfont}
\usepackage{tabularx}
\usepackage[export]{adjustbox}
\usepackage[normalem]{ulem}
\usepackage{arydshln}
\usepackage{caption}
\usepackage{wrapfig}

\title{Using Interleaved Ensemble Unlearning to Keep Backdoors at Bay for Finetuning Vision Transformers}
\author{
\hspace{-2pt}Zeyu Michael Li \\
Duke University\\
\texttt{zeyu.li030@duke.edu}}

\usepackage{xspace}

\newcommand{\poisonmodule}{poisoned module\xspace}
\newcommand{\robustmodule}{robust module\xspace}
\newcommand{\poisonmoduleSymbol}[1]{f_p(#1\,; \vtheta_p)}
\newcommand{\poisonmoduleSymbolShort}{f_p}
\newcommand{\robustmoduleSymbol}[1]{f_r(#1\,; \vtheta_r)}
\newcommand{\robustmoduleSymbolShort}{f_r}

\newcommand{\groundTruthOneHot}{\mathbf{y}}
\newcommand{\groundTruthOneHotPoison}{\mathbf{y}^{\hat{p}}}
\newcommand{\poisonModuleLogit}{\mathbf{\hat{y}}_p}
\newcommand{\robustModuleLogit}{\mathbf{\hat{y}}_r}

\newcommand{\logit}{\mathbf{\hat{y}}}
\newcommand{\weightMask}{m_{\vtheta_p}}  % renamed the text to logit masking
\newcommand{\numClasses}{N_c}

\newcommand{\poisonrate}{poisoning rate\xspace}
\newcommand{\confidencethresh}{confidence threshold\xspace}
\newcommand{\isolationratio}{isolation ratio\xspace}
\newcommand{\isolationratioSymbol}{r_{\text{isol}}}
\newcommand{\confidencethreshSymbol}{c_{\text{thresh}}}

\newcommand{\UnlearnSetSizeSymbol}{\hat{\alpha}}
\newcommand{\UnlearnLRSymbol}{lr^{\text{ul}}}
\newcommand{\TuneLRSymbol}{lr^{\text{tune}}}

\newcommand{\UnlearnSetSymbol}{\mathcal{D}^{\text{ul}}}

\newcommand{\FinetuneSetSymbol}{\mathcal{D}^{\text{tune}}}

\newcommand{\Image}{\mathbf{x}}
\newcommand{\potentialPoisonImage}{\Image^{\hat{p}}}
\newcommand{\ImageNoAug}{\Image_{\text{noAug}}}
\newcommand{\ImageYesAug}{\Image_{\text{yesAug}}}

\newcommand{\CELoss}[2]{\ell(#1, #2)}

\newcommand{\MyMethod}{IEU\xspace}

\begin{document}
\iclrfinalcopy

\maketitle
\begin{abstract}
Vision Transformers (ViTs) have become popular in computer vision tasks. Backdoor attacks, which trigger undesirable behaviours in models during inference, threaten ViTs' performance, particularly in security-sensitive tasks. Although backdoor defences have been developed for Convolutional Neural Networks (CNNs), they are less effective for ViTs, and defences tailored to ViTs are scarce. To address this, we present Interleaved Ensemble Unlearning (IEU), a method for finetuning clean ViTs on backdoored datasets. In stage 1, a shallow ViT is finetuned to have high confidence on backdoored data and low confidence on clean data. In stage 2, the shallow ViT acts as a ``gate'' to block potentially poisoned data from the defended ViT. This data is added to an unlearn set and asynchronously unlearned via gradient ascent. We demonstrate IEU's effectiveness on three datasets against 11 state-of-the-art backdoor attacks and show its versatility by applying it to different model architectures.
\end{abstract}

\section{Introduction}
\label{sec:introduction}
Vision Transformers (ViTs, \citet{vit-seminal-disovitskiy}) have emerged as a powerful alternative to Convolutional Neural Networks (CNNs) for a wide range of computer vision tasks. ViTs have achieved state-of-the-art performance in various downstream tasks such as image classification, object detection, and semantic segmentation \citep{vit-segmentation-survey,vit-obj-detection-survey}. However, the widespread deployment of ViTs have also raised concerns about their vulnerability to adversarial threats, particularly backdoor attacks, which typically modify images and/or labels in the training dataset to trigger attacker-controlled undesirable behaviour during inference \citep{badnets-seminal,backdoor-def-attn-block,vit-attack-catching-attn}. Backdoor attacks such as the BadNets attack in \citet{badnets-seminal} can compromise model behavior by embedding malicious triggers during training, leading to security risks in real-world applications. As ViTs become increasingly popular in security-sensitive domains such as autonomous driving and face recognition \citep{vit-autonomous-driving,vit-face-recognition}, it is important to understand these vulnerabilities and develop robust backdoor defences for ViTs.

ViTs are often pretrained using self-supervised learning (SSL) on large datasets and then finetuned to be deployed on specific tasks. Backdoor defences have been proposed to defend foundation models pretrained on large datasets and can either prevent backdoor injection during the SSL process or encourage removal of backdoors after pretraining \citep{vit-def-ssl1,vit-def-ssl2}; these thwart backdoor attacks that occur during pretraining, such as a practical real-world attack on web-scraped datasets in \citet{carlini-web-scale-backdoor} and an SSL-specific imperceptible attack \citet{ssl-attack-2024-0}. The finetuning process for adapting ViTs to downstream tasks using supervised learning is equally vulnerable to backdoor attacks. The rationale behind developing ViT-specific defences for finetuning are two-fold: \citet{vit-reliable-backdoors} shows that there are few defences specifically designed for ViTs for image classification in existing literature (\citet{vit-def-doan} and \citet{backdoor-def-attn-block} being notable examples of such defences); in addition, although existing defences designed for CNNs can defend ViTs after modifying the defence implementations, they still lead to high ASR and/or low CA when defending different flavours of ViTs (Tables 4 \& 5 in \citet{vit-reliable-backdoors}). 

\begin{figure}[t]
    \centering
    \includegraphics[width=0.99\linewidth]{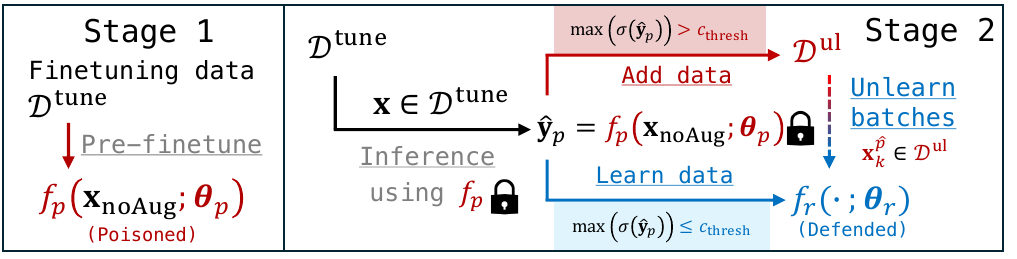}
    \vspace{-10pt}
    \caption{Overview of our defence, \textbf{\MyMethod}. The red \poisonmodule and blue \robustmodule are represented by $\poisonmoduleSymbolShort$ and $\robustmoduleSymbolShort$, respectively. Shaded boxes are conditions; underlined text represent actions. The lock icon indicates a frozen network; otherwise, it is trainable. The blue network is shielded from poisoned images by the red network and the blue network unlearns potentially poisoned data. The dynamic unlearning rate is not shown. Unaugmented images are used for $\poisonmoduleSymbolShort$ during both stages.}
    \vspace{-14pt}
    \label{fig:design-overview}
\end{figure}

To fill the gap, this work takes advantage of the \textit{highly modular architecture of the ViT} and propose a novel backdoor defence that uses an ensemble of two ViTs to perform interleaved unlearning on potentially poisoned data. Informed by the designs in \citet{denoised-poe} and \citet{abl-seminal}, we use a shallow ViT denoted by the ``\poisonmodule'' to defend the main ViT, which we call the ``\robustmodule''. Our design \textbf{\MyMethod} has two stages as shown in \Cref{fig:design-overview}. In \textbf{stage 1}, the \poisonmodule ($\poisonmoduleSymbolShort$), a shallow ViT, is tuned on the attacker-controlled finetuning data. Intuitively, shortcut learning \citep{geirhos-shortcut-seminal} leads $\poisonmoduleSymbolShort$ to learn shortcuts in the dataset, which are most prevalent in poisoned images. In addition, the simplicity of $\poisonmoduleSymbolShort$ discourages it from learning clean data that have fewer shortcuts. Therefore, the \poisonmodule is confident (where confidence is the maximum class probability $\max[\sigma(\poisonModuleLogit)]$ predicted by $\poisonmoduleSymbolShort$) when classifying poisoned data and not confident otherwise. In \textbf{stage 2}, images pass through the tuned $\poisonmoduleSymbolShort$, which either queues data onto the unlearn set or allows the defended main ViT (the \robustmodule) to learn data normally based on the \confidencethresh $\confidencethreshSymbol$. Whenever the unlearn set accumulates enough potentially poisoned data, a batch is unlearned by the \robustmodule using a dynamic unlearning rate. 
Instead of using a pre-determined unlearn set, our defence accumulates the unlearn set during stage 2. The benefits are two-fold: compared to using ABL's \citep{abl-seminal} method which isolates poisoned samples using the defended model, finetune-time unlearn set accumulation using $\poisonmoduleSymbolShort$ ensures that the \robustmodule learns as little poisoned data as possible; in addition, online accumulation of $\UnlearnSetSymbol$ is adaptive in the sense that the frequency of unlearning is high when more potentially poisoned images are encountered, quickly erasing the impact of a large number of poisoned data. In addition, we argue that core concepts developed in our method, namely applying interleaved ensemble unlearning, can defend other model architectures in image classification. Here are our main contributions:
\begin{itemize}[noitemsep,nolistsep,leftmargin=*]
    \item {We propose a new backdoor defence method, \MyMethod, \textit{specifically designed for defending a ViT} during finetuning. The defence uses an ensemble of two models to perform interleaved unlearning of high-confidence data using a dynamic unlearning rate. We show that our interleaved ensemble unlearning framework is \underline{successful without requiring high-precision isolation of poisoned data.}}
    \item {We argue that, in contrast to ABL, a fixed-sized $\UnlearnSetSymbol$ is not suitable for our design and demonstrate that Local Gradient Ascent (LGA) and loss Flooding presented in ABL \citep{abl-seminal} are orthogonal to our \poisonmodule (\Cref{sec:why-not-fix-sized-unlearn-set}). We show that the interleaved unlearning framework serves as a model-agnostic alternative to other unlearning-based defences (\Cref{tab:diff-vit-arch}). }
    \item {We empirically demonstrate that our design out-performs existing state-of-the-art defences on challenging datasets using 11 backdoor attacks by comparing to SOTA methods such as ABL and I-BAU \citep{abl-seminal,ibau}; Attack Success Rate (ASR) improved by \underline{33.83\%} and \underline{31.46\%} on average for TinyImageNet and CIFAR10, respectively, while maintaining high Clean Accuracy (CA). Our method requires no clean data and we perform extensive ablation studies.}
    \item {We explore potential points of failure of unlearning-based defence mechanisms to defend against \textit{weak} attacks where ``weakness'' corresponds to lower ASR. We propose potential solutions to address these failures. In our opinion, weak attacks are as insidious as powerful attacks.}
\end{itemize}

\section{Related work}
\label{sec:relatedwork}
\textbf{Backdoor Attacks}. The goals of backdoor attacks in image classification tasks are (a) to encourage a specific classification when the image is perturbed by an attacker-specified transformation and (b) to maintain normal performance when images without a backdoor trigger are classified \citep{badnets-seminal}. A great variety backdoor attacks of all flavours for both SSL \citep{ssl-attack-2024-0,ssl-attack-2024-1,ssl-attack-2022,ssl-attack-2023-embarrassing,ssl-attack-neil-gong-2022} and supervised learning \citep{backdoor-att-Blended,backdoor-att-Trojan,backdoor-att-wanet,backdoor-att-inv} have been proposed. There are three categories of backdoor attacks for supervised learning, which is the learning phase that this work defends: dirty-label attacks which includes visible and invisible attacks \citep{dirty-label-example-1,dirty-label-example-2,inv-example-1}, clean-label attacks which do not modify the label of backdoored images \citep{clean-label-example-1,clean-label-example-3-narcissus,clean-label-example-2-lcba}, and clean-image attacks which only modify data labels during learning \citep{backdoor-att-clean-image,clean-image-example-1}. Authors have also developed \textit{ViT-specific} backdoor attacks. For example, \citet{vit-specific-trojvit} inserts a Trojan into a ViT checkpoint, while \citet{vit-specific-dbia} modifies the finetuning procedure by using an attacker-specified loss function. 

\textbf{Backdoor Defences}. The defender's goal is to ensure that backdoor images do not trigger attacker-specified model behaviour whilst maintaining high CA. A popular class of defences is model reconstruction where defenders cleanse poisoned models of backdoors. Early works \citep{backdoor-def-cleanse-1,backdoor-def-cleanse-2} in this area such as the fine-pruning defence prunes neurons that are dormant on clean images and tunes on clean data. Later works that aim to remove backdoor neurons include Neural Attention Distillation (NAD, \citet{backdoor-def-cleanse-4-nad}), Adversarial Neuron Pruning (ANP, \citet{backdoor-def-cleanse-5-anp}) Adversarial Weight Masking (AWM, \citet{backdoor-def-cleanse-3-awm}), Shapley-estimation based few-shot defence \citep{backdoor-def-cleanse-6-shapley}, and Reconstructive Neuron Pruning (RNP, \citet{rnp}). Another such cleansing defence, I-BAU \citep{ibau} connects the two optimisation problems in the minimax formulation of backdoor removal using an implicit hypergradient. Authors invariably design insightful methods for recovering clean models after backdoor injection. Certified backdoor defences have also been developed \citep{backdoor-def-certified}. Another broad class of defences involves reconstructing the trigger in order to unlearn backdoor images. Notable examples include Neural Cleanse \citep{backdoor-def-trigger-recon-1-nc}, DeepInspect \citep{backdoor-def-trigger-recon-2-DeepInspect} which checks for signs of backdooring without a reserved clean set, and BTI-DBF \citep{backdoor-def-trigger-recon-3-bti-dbf} which decouples benign features for backdoor trigger inversion. Mitigating backdoor attacks during tuning \citep{backdoor-def-train-suppress-2,backdoor-def-t-and-r,backdoor-def-train-suppress-1} is also popular. Our method is closely related to Anti-Backdoor Learning (ABL, \citet{abl-seminal}), which isolates poisoned data and unlearns them after training. 

\textbf{ViT-Specific Backdoor Defences}. Few backdoor defences are specifically designed for defending ViTs during tuning \citep{vit-reliable-backdoors} and existing defences that have CNNs in mind perform worse on ViTs. Two notable defences are \citet{vit-def-doan} where backdoor images are identified using patch-processing, and \citet{backdoor-def-attn-block} which is a test-time defence that uses GradRollout (\citet{GradRollout}, an interpretation method for ViTs) to block high-attention patches in images.

\section{Method}
\label{sec:method}
In this section, we first define our threat model and describe \MyMethod in detail. We conclude this section by briefly exploring the drawbacks of using a fixed-size unlearn set $\UnlearnSetSymbol$.

\textbf{Threat model}. We focus on finetuning ViTs for image classification tasks and assume that the pretrained model checkpoint initially given to the defender is not benign. We follow the threat model of \citet{abl-seminal}. We assume that the finetuning procedure is controlled by the defender, which means attacks that modify finetuning loss or the model's gradient \citep{vit-specific-dbia,blind-backdoor-loss-modified} are out of scope. On the other hand, finetuning data is gathered from untrusted sources and may contain backdoor data. The attacker knows the model architecture and the pretrained checkpoint's parameter values, and may poison the finetuning dataset by modifying images and/or labels. The defender aims to tune a benign checkpoint for downstream tasks using $\FinetuneSetSymbol$ and does not know the distribution/proportion of backdoor data in the attacker-supplied $\FinetuneSetSymbol$.

\textbf{Notations}. The finetuning set $\FinetuneSetSymbol$ \textit{may} contain an unknown proportion of backdoor samples; this proportion (i.e., \poisonrate) is denoted by $\alpha$. The defender unlearns data in the unlearn set $\UnlearnSetSymbol$, whose size as a fraction of $\FinetuneSetSymbol$ is defined as $\UnlearnSetSizeSymbol = |\UnlearnSetSymbol|\div|\FinetuneSetSymbol|$. The two sub-networks in the ensemble are the \poisonmodule and \robustmodule, denoted by $\poisonmoduleSymbolShort$ and $\robustmoduleSymbolShort$, respectively. Data points $(\Image,\groundTruthOneHot)\in\FinetuneSetSymbol$, which consist of unaugmented ($\ImageNoAug$) and augmented ($\ImageYesAug$) views of the original image (as in ``data augmentation''), and potentially poisoned data points $(\potentialPoisonImage, \groundTruthOneHotPoison) \in\UnlearnSetSymbol$ are used to finetune and defend $\robustmoduleSymbolShort$, respectively. For simplicity, we use $\Image$ to denote images when data augmentation is not relevant. The logits produced by the two modules are referred to as $\poisonModuleLogit=\poisonmoduleSymbol{\Image}$ and $\robustModuleLogit=\robustmoduleSymbol{\Image}$, where $\vtheta_p,\vtheta_r$ are the potentially tunable parameters of $\poisonmoduleSymbolShort$ and $\robustmoduleSymbolShort$, respectively. The two logits vectors $\poisonModuleLogit$ and $\robustModuleLogit$ combine to form the logits vector $\logit$ based on $\weightMask$ (\Eqref{eqn:weight-masking}). We use $\sigma(\cdot)$ and $\CELoss{\cdot}{\cdot}$ to represent the softmax function and the cross-entropy loss, respectively. The confidence threshold $0<\confidencethreshSymbol<1$ determines whether an image is asynchronously unlearned or immediately learned. The number of classes in $\FinetuneSetSymbol$ is denoted by $\numClasses$ for CIFAR10, GTSRB, and TinyImageNet, respectively. The learning rates used to finetune $\robustmoduleSymbolShort$ and unlearn $\potentialPoisonImage$ are $\TuneLRSymbol$ and $\UnlearnLRSymbol$, respectively.

\subsection{Interleaved Ensemble Unlearning (\textbf{\MyMethod})}
\textbf{Overview.} \Cref{fig:design-overview} summarises our method, which has two stages. During stage 1, the \poisonmodule $\poisonmoduleSymbol{\cdot}$ is pre-finetuned using finetuning data $\FinetuneSetSymbol$. During stage 2, $\poisonmoduleSymbolShort$ is used to determine whether incoming data is learned by the \robustmodule $\robustmoduleSymbol{\cdot}$ or added to $\UnlearnSetSymbol$ for asynchronous unlearning based on the unlearn rate $\UnlearnLRSymbol$ in \Eqref{eqn:unlearn-rate}. 

\textbf{Stage 1: Pre-finetuning the \poisonmodule $\poisonmoduleSymbolShort$}. This step applies vanilla tuning (for hyperparameters see \Cref{tab:defence-param} in \Cref{sec:appendix-defence-details}) using $\FinetuneSetSymbol$ on $\poisonmoduleSymbol{\cdot}$, where $\vtheta_p$ is initialised as the first few layers of a pretrained checkpoint. Stage 1 solves the following optimisation problem: $\min_{\vtheta_p} \E_{\ImageNoAug \sim \FinetuneSetSymbol}[\CELoss{\poisonmoduleSymbol{\ImageNoAug}}{\groundTruthOneHot}]$, where $\CELoss{\cdot}{\cdot}$ is the cross entropy loss, $\groundTruthOneHot$ is the one-hot ground truth vector, and $\poisonmoduleSymbolShort$ is the \poisonmodule. As explained in \Cref{sec:introduction}, the intuition of overfitting $\poisonmoduleSymbolShort$ on poisoned data is based on shortcut learning \citep{geirhos-shortcut-seminal}. These shortcuts are found overwhelmingly in unaugmented backdoored images, where the attacker-specified trigger acts as an easily identifiable artifact that causes $\poisonmoduleSymbolShort$ to easily learn the connection between the trigger and attacker-specified label. Therefore, the tuned $\poisonmoduleSymbolShort$ is likely to be confident when predicting images with the trigger. The \poisonmodule $\poisonmoduleSymbolShort$ is designed to be complex enough to learn shortcuts and shallow enough to avoid learning much from benign data. The goal is for $\max(\sigma(\poisonmoduleSymbol{\ImageNoAug}))$ to be small for clean images and large for backdoor images after tuning in stage 1. 

\textbf{Stage 2: Finetuning the \robustmodule $\robustmoduleSymbolShort$}. This stage applies defended finetuning on $\robustmoduleSymbol{\cdot}$ which optimises the objectives in Equations \ref{eqn:stage2-optim} and \ref{eqn:stage2-optim-unlearn}; $\vtheta_r$ is initialised as a pretrained checkpoint. During this stage, $\poisonmoduleSymbolShort$ is frozen and only $\robustmoduleSymbolShort$ is tuned. The logits $\poisonModuleLogit=\poisonmoduleSymbol{\ImageNoAug}$ for the unaugmented view of each image is produced by the \poisonmodule in order to compute maximum class probability $\max(\sigma(\poisonModuleLogit))$, which is then compared to $\confidencethreshSymbol$ to determine whether the data point should be learned by $\robustmoduleSymbolShort$ or added onto $\UnlearnSetSymbol$. If $\max(\sigma(\poisonModuleLogit))$ is above the \confidencethresh $\confidencethreshSymbol$, the data point is added to $\UnlearnSetSymbol$. Otherwise, $\robustmoduleSymbolShort$ learns the augmented views as would happen in regular finetuning. To prevent potentially poisoned data from being learned, we apply \textit{logit masking} on the output logits of $\robustmoduleSymbolShort$ and $\poisonmoduleSymbolShort$ as shown in \Eqref{eqn:weight-masking}
\begin{equation}
\label{eqn:weight-masking}
\weightMask= \1_{x<\confidencethreshSymbol}(\max(\sigma(\poisonmoduleSymbol{\ImageNoAug}))),\text{ where  } \logit = \poisonModuleLogit (1-\weightMask) + \robustModuleLogit \weightMask
\end{equation}
where $\weightMask$ is the binary logit mask, $\1_{x<\confidencethreshSymbol}(x)$ is the indicator function, $\groundTruthOneHot$ is the ground truth, $\logit$ is the logits vector, and $\poisonModuleLogit=\poisonmoduleSymbol{\ImageNoAug},\,\robustModuleLogit=\robustmoduleSymbol{\ImageYesAug}$ are the logits produced by $\poisonmoduleSymbolShort$ and $\robustmoduleSymbolShort$, respectively. When $\ImageNoAug$ is detected by $\poisonmoduleSymbolShort$ as a potentially poisoned image, the logits for optimising the objective in \Eqref{eqn:stage2-optim} come from $\poisonmoduleSymbolShort$; otherwise, $\logit = \robustModuleLogit$ for optimisation. In other words, optimising the objective in \Eqref{eqn:stage2-optim} requires both $\poisonmoduleSymbolShort$ and $\robustmoduleSymbolShort$ to contribute to the logits. 
\begin{equation}\label{eqn:stage2-optim}
\min_{\vtheta_r} \E_{\Image\sim\FinetuneSetSymbol}[\CELoss{\logit}{\groundTruthOneHot}]
\end{equation}
The unlearn set $\UnlearnSetSymbol$ accumulates data until it has enough data for one batch containing potentially poisoned images $(\potentialPoisonImage, \groundTruthOneHotPoison)$, which is then unlearned by $\poisonmoduleSymbolShort$ during finetuning. Unlike the continuously decaying learning rate $\TuneLRSymbol$ used for normal finetuning, the unlearning rate $\UnlearnLRSymbol$ doesn't depend on just the decay schedule. Given $\TuneLRSymbol$ which follows the cosine annealing decay schedule, the $(k-1)^\text{th}$ batch with potentially poisoned images $(\potentialPoisonImage_{k-1}, \groundTruthOneHotPoison_{k-1})$, the number of classes in the dataset $\numClasses$, and the \robustmodule $\robustmoduleSymbol{\cdot}$, the dynamic unlearning rate for the current batch $(\potentialPoisonImage_{k}, \groundTruthOneHotPoison_{k})$ is defined in \Eqref{eqn:unlearn-rate} and can be viewed as a function of cross entropy loss of the previous batch of potentially poisoned images $\CELoss{\robustmoduleSymbol{\potentialPoisonImage_{k-1}}}{\groundTruthOneHotPoison_{k-1}}$.
\begin{equation}
\label{eqn:unlearn-rate}
\resizebox{0.94\textwidth}{!}{
    $\UnlearnLRSymbol_k = \TuneLRSymbol\cdot \left(\1_{k>0}(k)\cdot\max\left[6 - \exp\left[-\left(\ln(\numClasses) - \CELoss{\robustmoduleSymbol{\potentialPoisonImage_{k-1}}}{\groundTruthOneHotPoison_{k-1}}\div\sqrt{2}\right)\right], 0.2\right]+\1_{k=0}(k)\right) $
    }
\end{equation}
The two indicator functions ensure $\UnlearnLRSymbol_k = \TuneLRSymbol$ when $k=0$, which occurs at an epoch's beginning. The term that scales $\TuneLRSymbol$ in \Eqref{eqn:unlearn-rate} is an exponentially decreasing function with respect to increasing loss $\CELoss{\robustmoduleSymbol{\potentialPoisonImage_{k-1}}}{\groundTruthOneHotPoison_{k-1}}$, causing $\UnlearnLRSymbol_k$ to be large when the previous batch $\potentialPoisonImage_{k-1}$ produces low cross entropy loss on $\robustmoduleSymbolShort$. This keeps the backdoor from being learned by $\robustmoduleSymbolShort$. Although it is shown in \Cref{tab:const-unlearn-rate} that using $\UnlearnLRSymbol_k=c\cdot\TuneLRSymbol$ for some $c\in \R^+$ performs better than using \Eqref{eqn:unlearn-rate}, one benefit of defining $\UnlearnLRSymbol_k$ using a fixed function is that $\UnlearnLRSymbol_k$ is no longer a hyperparameter that needs to be tuned. In addition to solving \Eqref{eqn:stage2-optim}, finetuning during stage 2 also optimises for the objective in \Eqref{eqn:stage2-optim-unlearn}, which is implemented using gradient ascent performed on $\robustmoduleSymbolShort$ given $(\potentialPoisonImage,\groundTruthOneHotPoison)\in\UnlearnSetSymbol$. See \Cref{alg:stage2} in \Cref{sec:algorithms} for a precise description of stage 2. 
\begin{equation}
\label{eqn:stage2-optim-unlearn}
    \max_{\vtheta_r} \E_{\potentialPoisonImage\sim\UnlearnSetSymbol}[\CELoss{\robustmoduleSymbol{\potentialPoisonImage}}{\groundTruthOneHotPoison}]
\end{equation}

\subsection{Why Not a Fixed-Sized Unlearn Set?}
\label{sec:why-not-fix-sized-unlearn-set}
In this subsection, we argue that our method of using $\poisonmoduleSymbolShort$ to isolate poisoned data is orthogonal to similar methods used in ABL \citep{abl-seminal} and is more suitable for our \MyMethod. Authors in ABL isolates a fixed fraction (called ``$\isolationratioSymbol$'') of the tuning set where $0\leq \isolationratioSymbol\leq 1$ (they used $\isolationratioSymbol=0.01$). In their method, images whose cross entropy loss rank amongst the lowest $\isolationratioSymbol$ fraction of $\FinetuneSetSymbol$ is collected to form $\UnlearnSetSymbol$ of size $\UnlearnSetSizeSymbol=\isolationratioSymbol$. ABL uses techniques such as Local Gradient Ascent (LGA) or loss Flooding \textit{on the defended model} to encourage poisoned images to have low loss. Compared to using $\isolationratioSymbol$ (ABL), there are two main advantages for using $\confidencethreshSymbol$ (our method) to produce the unlearn set in our method: (a) the effectiveness (FPR or FNR) of our isolation method is not significantly affected by the value of the \poisonrate $\alpha$, which is unknown to the defender (\Cref{tab:poison_rate_variable}), and (b) the unlearn set size varies as $\alpha$ varies, which increases defence success using \MyMethod since a high proportion of poisoned data should be added to $\UnlearnSetSymbol$ for low ASR and high CA ($\UnlearnSetSizeSymbol_i\div\alpha\in\{0.9, 1.0\}$ in \Cref{tab:unlearning_set_size_variable}).

\begin{table}[!h]
    \caption{Performance of the two methods when evaluated on detecting poisoned finetuning data (CIFAR10). Five \textit{\poisonmodule} ($\poisonmoduleSymbolShort$) instances are pre-finetuned for 10 epochs at $2\cdot 10^{-4}$ learning rate with different \poisonrate $\alpha$; $\confidencethreshSymbol$ and $\isolationratioSymbol$ are fixed at $0.95$ and $0.1$, respectively. Each cell shows the FPR/FNR values as percentages (``positive'' means ``poisoned''). }
    \label{tab:poison_rate_variable}
    \vspace{-9pt}
    \centering
    \scalebox{0.9}{
    \begin{tabular}{ll ccccc}
        \toprule
        Attack & Selection Method & $\alpha=0.02$ & 0.05 & 0.10 & 0.15 & 0.20 \\
        \midrule
        \multirow[m]{2}{*}{BadNets-white} & $\isolationratioSymbol$ & 8.68/25.30 & 6.42/22.04 & 0.80/7.22 & 0.03/33.49 & 0.00/50.00 \\
         & $\confidencethreshSymbol$ & 9.16/24.90 & 5.67/22.84 & 5.62/5.30 & 6.09/10.25 & 4.71/5.17 \\
        \cline{1-7}
        \multirow[m]{2}{*}{ISSBA} & $\isolationratioSymbol$ & 8.59/21.00 & 5.60/6.44 & 0.62/5.54 & 0.00/33.33 & 0.00/50.00 \\
         & $\confidencethreshSymbol$ & 6.27/25.40 & 6.79/5.76 & 5.55/2.58 & 6.51/1.48 & 5.68/0.48 \\
        \cline{1-7}
        \bottomrule
    \end{tabular}
    }
\end{table}

\begin{table}[!h]
    \centering
    \caption{Performance of models that are defended during stage 2 using \MyMethod with hand-crafted unlearn sets $\UnlearnSetSymbol_i$ of varying sizes ($\UnlearnSetSizeSymbol_i$) as a fraction of the \textit{original} finetune set $\FinetuneSetSymbol$. The \poisonrate is fixed at $\alpha=0.1$ and the sizes of $\UnlearnSetSymbol_i$ as a fraction of $\FinetuneSetSymbol$ are $\UnlearnSetSizeSymbol_i\in(0.01, 0.02, 0.05, 0.1, 0.2)$. A hand-crafted unlearn set $\UnlearnSetSymbol_i$ consists entirely of poisoned data if $\UnlearnSetSizeSymbol_i\div\alpha \leq 1$ and includes all poisoned data if $\UnlearnSetSizeSymbol_i\div\alpha \geq 1$. All values here are expressed in percentages. }
    \label{tab:unlearning_set_size_variable}
    \vspace{-9pt}
    \scalebox{0.68}{
    \begin{tabular}{llcccccccccccc}
        \toprule
         \multicolumn{2}{r}{Size ratio ($\UnlearnSetSizeSymbol_i\div\alpha$):}  & \multicolumn{2}{c}{0.1} & \multicolumn{2}{c}{0.2} & \multicolumn{2}{c}{0.5} & \multicolumn{2}{c}{0.9} & \multicolumn{2}{c}{1.0} & \multicolumn{2}{c}{2.0} \\
         \cmidrule(lr){3-4} \cmidrule(lr){5-6} \cmidrule(lr){7-8} \cmidrule(lr){9-10} \cmidrule(lr){11-12}  \cmidrule(lr){13-14}
        Dataset & Attack & ASR & CA & ASR & CA & ASR & CA & ASR & CA & ASR & CA & ASR & CA \\
        \midrule
        \multirow[c]{3}{*}{CIFAR10} & BadNets-white & 10.09 & 98.18 & 9.89 & 97.58 & 6.41 & 96.30 & 0.88 & 97.94 & 0.88 & 98.28 & 0.67 & 97.83 \\
         & ISSBA & 100.00 & 98.31 & 100.00 & 98.18 & 0.00 & 96.94 & 0.00 & 97.99 & 0.00 & 98.23 & 0.00 & 98.16 \\
         & Smooth & 95.32 & 98.24 & 82.28 & 97.93 & 0.06 & 97.00 & 0.77 & 97.98 & 0.94 & 98.24 & 0.11 & 97.90 \\
        \cline{1-14}
        \multirow[c]{3}{*}{TinyImageNet} & BadNets-white & 0.32 & 61.95 & 0.01 & 57.89 & 0.00 & 54.03 & 0.00 & 63.43 & 0.00 & 65.95 & 0.00 & 37.18 \\
         & ISSBA & 57.84 & 64.94 & 0.15 & 58.70 & 0.00 & 22.50 & 0.00 & 38.63 & 0.16 & 46.97 & 0.05 & 16.37 \\
         & Smooth & 93.51 & 68.37 & 77.33 & 64.84 & 0.00 & 55.93 & 0.00 & 61.95 & 0.01 & 66.28 & 0.00 & 38.86 \\
        \cline{1-14}
        \bottomrule
    \end{tabular}
    }
\end{table}
\Cref{tab:poison_rate_variable} shows that the FPR/FNR are similar for both $\isolationratioSymbol$ and $\confidencethreshSymbol$ at low \poisonrate. However, as $\alpha$ increases, using $\isolationratioSymbol$ causes more poisoned data to be left out of $\UnlearnSetSymbol$. For example, at $\alpha=0.2$ and $\isolationratioSymbol=0.1$ (meaning $\UnlearnSetSizeSymbol_i\div\alpha=0.1\div0.2=0.5$), the FNR is $50\%$. This results in instability during defence and worse performance as shown in \Cref{tab:unlearning_set_size_variable} (column 0.5) since a large fraction of poisoned data is not in $\UnlearnSetSymbol$. As less poisoned data is included in $\UnlearnSetSymbol$, our defence becomes less effective with ASR increasing and CA decreasing (\Cref{tab:unlearning_set_size_variable}). Since using a fixed $\isolationratioSymbol$ leaves many poisoned images outside of $\UnlearnSetSymbol$ when $\alpha>\isolationratioSymbol$, we use $\confidencethreshSymbol$ to select a variable-sized $\UnlearnSetSymbol$.

We show in \Cref{tab:lga-flooding-no-method} (\Cref{sec:appendix-ablation}) that our shallow $\poisonmoduleSymbolShort$ is not compatible with LGA/Flooding for isolating poisoned data; they are meant to be orthogonal methods.

\section{Experiments}
\label{sec:experiments}
See \Cref{sec:more-implementation-details} for more details regarding baselines, attacks, datasets and defence parameters.

\textbf{Baselines}. We use three baseline methods for comparison with our defence. Specifcally, we compare against I-BAU \citep{ibau}, ABL \citep{abl-seminal}, and AttnBlock \citep{backdoor-def-attn-block} which is a ViT-specific defence. We report results for AttnBlock in \Cref{sec:appendix-defence-details} due to high ASR. I-BAU and ABL are state-of-the-art general defences not specifically designed for ViTs; authors in \citep{backdoor-def-t-and-r} suggest that I-BAU is the most competitive baseline compared to others. We attempt and fail to reproduce the RNP defence \citep{rnp} for ViTs despite following recommendations in \citet{vit-reliable-backdoors} to mask features of linear layers instead of those of norm layers.

\begin{table}[!b]
    \centering
    \caption{Performance of \MyMethod compared to no defence, ABL \citep{abl-seminal}, I-BAU \citep{ibau} given as percentages. Averages of each column are given in the last row for that dataset and best/second-best values are bolded/underlined. See \Cref{sec:appendix-defence-details} for AttnBlock results. }
    \label{tab:all-results}
    \vspace{-9pt}

    \scalebox{0.92}{
    \begin{tabular}{llcccccccccc}
\toprule
 \multirow[c]{2}{*}{Dataset} & \multirow[c]{2}{*}{Attack} & \multicolumn{2}{c}{No Defence} & \multicolumn{2}{c}{I-BAU} & \multicolumn{2}{c}{ABL} & \multicolumn{2}{c}{\textbf{\MyMethod (Ours)}} \\
 \cmidrule(lr){3-4} \cmidrule(lr){5-6} \cmidrule(lr){7-8} \cmidrule(lr){9-10}
 &  & ASR & CA & ASR & CA & ASR & CA & ASR & CA \\
\midrule
\multirow[c]{12}{*}{CIFAR10} & BadNets-white & 97.51 & 98.36 & 10.12 & 95.86 & \underline{9.7} & \underline{98.11} & \textbf{0.96} & \textbf{98.19} \\
 & BadNets-pattern & 100.0 & 98.23 & \underline{91.84} & 92.16 & 100.0 & \underline{97.4} & \textbf{0.0} & \textbf{98.22} \\
 & ISSBA & 100.0 & 98.13 & \underline{9.46} & \underline{92.96} & 100.0 & 37.62 & \textbf{0.33} & \textbf{98.35} \\
 & BATT & 100.0 & 98.28 & \underline{21.68} & 95.86 & 90.05 & \underline{97.94} & \textbf{0.02} & \textbf{98.23} \\
 & Blended & 100.0 & 98.39 & \underline{21.0} & 93.92 & 25.04 & \underline{97.82} & \textbf{0.0} & \textbf{98.27} \\
 & Trojan-WM & 99.99 & 98.32 & \underline{79.94} & 94.16 & 99.91 & \underline{97.97} & \textbf{0.0} & \textbf{98.15} \\
 & Trojan-SQ & 99.7 & 98.31 & \underline{68.26} & 94.62 & 99.62 & \textbf{98.27} & \textbf{0.04} & \underline{98.22} \\
 & Smooth & 99.76 & 98.24 & 16.26 & 94.0 & \underline{9.31} & \underline{97.12} & \textbf{0.09} & \textbf{97.77} \\
 & l0-inv & 100.0 & 98.34 & 10.28 & 93.24 & \underline{0.04} & \underline{97.12} & \textbf{0.0} & \textbf{98.19} \\
 & l2-inv & 99.98 & 98.41 & 9.98 & 93.4 & \underline{8.62} & \underline{98.1} & \textbf{0.44} & \textbf{98.24} \\
 & SIG & 98.49 & 88.75 & \underline{9.16} & \textbf{94.38} & 97.94 & \underline{88.44} & \textbf{0.0} & 87.67 \\

 & \textbf{Average} & 99.58 & 97.43 & \underline{31.63} & \underline{94.05} & 58.2 & 91.45 & \textbf{0.17} & \textbf{97.23} \\
 \cline{1-10}
\multirow[c]{12}{*}{GTSRB} & BadNets-white & 95.7 & 95.63 & 5.48 & \textbf{99.1} & \underline{4.09} & \underline{92.83} & \textbf{0.93} & 82.07 \\
 & BadNets-pattern & 100.0 & 96.44 & 5.46 & \textbf{98.17} & \underline{4.09} & \underline{93.76} & \textbf{0.0} & 88.53 \\
 & ISSBA & 99.99 & 95.91 & \underline{5.48} & \textbf{99.48} & \textbf{3.63} & 93.15 & \cellcolor{red!10}100.0 & \underline{93.52} \\
 & BATT & 100.0 & 96.18 & 6.08 & \textbf{99.58} & \underline{2.29} & 92.79 & \textbf{0.03} & \underline{93.47} \\
 & Blended & 100.0 & 96.95 & 11.24 & \textbf{97.75} & \textbf{0.0} & \underline{92.95} & \underline{8.76} & 81.2 \\
 & Trojan-WM & 100.0 & 92.75 & 5.26 & \textbf{98.11} & \textbf{0.0} & 75.16 & \underline{3.12} & \underline{86.54} \\
 & Trojan-SQ & 99.85 & 94.91 & \underline{5.48} & \textbf{99.61} & 99.6 & \underline{81.99} & \textbf{0.07} & 77.29 \\
 & Smooth & 99.79 & 96.29 & 27.5 & \textbf{99.74} & \textbf{3.46} & \underline{92.36} & \underline{6.15} & 84.52 \\
 & l0-inv & 100.0 & 96.76 & \underline{5.44} & \textbf{99.7} & 100.0 & \underline{94.96} & \textbf{0.0} & 87.48 \\
 & l2-inv & 100.0 & 93.93 & \textbf{9.06} & \textbf{99.71} & 77.79 & \underline{96.03} & \underline{12.85} & 87.5 \\
 & SIG & 99.52 & 91.41 & \underline{2.92} & 38.32 & 95.53 & \textbf{83.49} & \textbf{0.21} & \underline{72.07} \\

 & \textbf{Average} & 99.53 & 95.2 & \textbf{8.13} & \textbf{93.57} & 35.5 & \underline{89.95} & \underline{12.01} & 84.93 \\
 \cline{1-10}
\multirow[c]{12}{*}{TinyImageNet} & BadNets-white & 98.51 & 61.46 & 0.48 & 51.06 & \underline{0.24} & \underline{59.19} & \textbf{0.12} & \textbf{66.35} \\
 & BadNets-pattern & 100.0 & 62.72 & 75.72 & 55.04 & \underline{0.25} & \underline{60.59} & \textbf{0.0} & \textbf{66.62} \\
 & ISSBA & 99.62 & 63.1 & 87.18 & 0.66 & \underline{0.08} & \textbf{57.56} & \textbf{0.05} & \underline{40.6} \\
 & BATT & 99.98 & 66.66 & 89.8 & 60.16 & \textbf{0.02} & \underline{62.36} & \underline{3.07} & \textbf{64.15} \\
 & Blended & 100.0 & 70.21 & 65.22 & 61.38 & \textbf{0.0} & \underline{64.06} & \textbf{0.0} & \textbf{66.41} \\
 & Trojan-WM & 99.96 & 69.89 & \underline{90.68} & 60.9 & 99.31 & \textbf{68.92} & \textbf{0.0} & \underline{67.33} \\
 & Trojan-SQ & 99.79 & 63.56 & \underline{97.5} & 57.46 & 99.74 & \underline{63.04} & \textbf{0.0} & \textbf{67.28} \\
 & Smooth & 99.35 & 68.58 & 4.62 & 60.68 & \textbf{0.03} & \underline{61.74} & \underline{0.17} & \textbf{66.03} \\
 & l0-inv & 100.0 & 63.14 & \underline{99.76} & \underline{54.48} & 99.99 & 44.29 & \textbf{0.0} & \textbf{65.89} \\
 & l2-inv & 99.82 & 65.43 & 0.44 & \underline{59.32} & \underline{0.04} & 57.92 & \textbf{0.01} & \textbf{64.5} \\
 & SIG & 67.99 & 71.65 & \underline{19.04} & \underline{63.58} & 83.67 & 59.04 & \textbf{7.77} & \textbf{71.71} \\

 & \textbf{Average} & 96.82 & 66.04 & 57.31 & 53.16 & \underline{34.85} & \underline{59.88} & \textbf{1.02} & \textbf{64.26} \\
 \cline{1-10}
\bottomrule
    \end{tabular}
    }
\end{table}

\textbf{Attacks}. We evaluate the performance of our design on 11 backdoor attacks. Specifically, we consider 9 out of 10 attacks in \citet{backdoor-def-t-and-r}: BadNets-white (white lower-right corner), BadNets-pattern (grid pattern in lower-right corner) \citep{badnets-seminal}, Blended \citep{backdoor-att-Blended}, l0-inv, l2-inv \citep{backdoor-att-inv}, Smooth \citep{backdoor-att-Smooth}, Trojan-SQ, Trojan-WM \citep{backdoor-att-Trojan}, and a clean label attack, SIG \citep{backdoor-att-sig-clean-label}. In addition, we consider the sample-specific invisible attack ISSBA \citep{backdoor-att-issba} and an image transformation-based attack BATT \citep{backdoor-att-batt}. Please refer to \Cref{sec:appendix-attack-details} for visualisations of backdoored images. Although ViT-specific backdoor attacks exist in literature \citep{vit-specific-trojvit, vit-specific-dbia}, we did not include these attacks due to their focus on inference-time attacks. Moreover, they use threat models that are incompatible with ours. For example, \citet{vit-specific-trojvit} injects a trigger at inference-time (which does not concern finetuning), while \citet{vit-specific-dbia} modifies the finetuning procedure (which is controlled by the defender in our threat model) by using an attacker-specified loss function. 

\textbf{Datasets and default parameters}. We used three datasets to evaluate our defence (CIFAR10 \citet{cifar10-dataset}, GTSRB \citet{gtsrb-dataset}, and TinyImageNet\footnote{Used in Stanford's course CS231N. Download: \scriptsize\url{http://cs231n.stanford.edu/tiny-imagenet-200.zip}} \citet{imagenet-dataset}). Defence parameters are shown in \Cref{sec:appendix-defence-details}. 

\subsection{Main Results}
We show the results of our \MyMethod compared to other baselines in \Cref{tab:all-results}. Our method's ASR out-performs I-BAU by $31.46$ percentage points (pp) on CIFAR10 and out-performs ABL by $33.83$pp on TinyImageNet. In addition, our \MyMethod's CA for CIFAR10 and TinyImageNet are generally better than the corresponding values of the baselines. Our method has the lowest ASR in all attacks and 9 out of 11 attacks in CIFAR10 and TinyImageNet, respectively. Moreover, our method produces the highest CA for 9 out of 11 attacks for both CIFAR10 and TinyImageNet. We explore the limitations of \MyMethod in \Cref{sec:discussion} for weaker attacks and for the GTSRB dataset. Note that I-BAU uses the highest amount of GPU memory (39 GB on an NVIDIA A100) when compared to ABL and \MyMethod ($\leq20$ GB).

\subsection{Ablation Study on Hyperparameters}
\textbf{The importance of logit masking} is shown in \Cref{tab:no-weight-masking}, which demonstrates performance degradation of \MyMethod when logit masking is not used. Without logit masking, $\robustmoduleSymbolShort$ both learns and unlearns $(\potentialPoisonImage, \groundTruthOneHotPoison) \in \UnlearnSetSymbol$, which by default is not learned in \MyMethod. If the \robustmodule performs poorly on potentially poisoned data because of asynchronous unlearning, the absence of logit masking allows the model to relearn the poisoned data during parameter updates for finetuning. Therefore, the model repeatedly learns and unlearns the same data, resulting in low performance on non-poisoned data. This reasoning also guides our decision to use $\poisonmoduleSymbolShort$ instead of $\robustmoduleSymbolShort$ to isolate $\potentialPoisonImage$ for Interleaved Unlearning.
\begin{table}[!h]
    \caption{Performance of \MyMethod without applying logit masking during finetuning. Leaving out logit masking means that $\logit=\robustModuleLogit$ is used instead of using \Eqref{eqn:weight-masking}. All values given in percentages. }
    \label{tab:no-weight-masking}
    \centering
    \vspace{-9pt}
    \scalebox{0.9}{
    \begin{tabular}{lcccccccc}
        \toprule
          \multirow[c]{2}{*}{Dataset} & \multicolumn{2}{c}{BATT} & \multicolumn{2}{c}{BadNets-white} & \multicolumn{2}{c}{ISSBA} & \multicolumn{2}{c}{Smooth} \\
         \cmidrule(lr){2-3} \cmidrule(lr){4-5} \cmidrule(lr){6-7} \cmidrule(lr){8-9}
         & ASR & CA & ASR & CA & ASR & CA & ASR & CA \\
        \midrule
        CIFAR10 & 0.00 & 79.75 & 0.61 & 81.49 & 0.00 & 83.42 & 0.00 & 64.06 \\
        TinyImageNet & 0.36 & 56.12 & 0.01 & 28.85 & 0.00 & 11.98 & 0.00 & 26.49 \\
        \bottomrule
    \end{tabular}
    }
\end{table}
\begin{table}[!h]
    \centering
    \caption{Performance of \MyMethod when finetuned using $\FinetuneSetSymbol$ with different \poisonrate values using CIFAR10, $\alpha=\{0.02, 0.05, 0.1, 0.15, 0.2\}$. All values given in percentages. }
    \vspace{-9pt}
    \label{tab:diff-poi-defended}
    \scalebox{0.9}{\begin{tabular}{lcccccccccc}
        \toprule
         \multirow[c]{2}{*}{Attack} & \multicolumn{2}{c}{$\alpha=0.02$} & \multicolumn{2}{c}{0.05} & \multicolumn{2}{c}{0.10} & \multicolumn{2}{c}{0.15} & \multicolumn{2}{c}{0.20} \\
         \cmidrule(lr){2-3} \cmidrule(lr){4-5} \cmidrule(lr){6-7} \cmidrule(lr){8-9} \cmidrule(lr){10-11}
         & ASR & CA & ASR & CA & ASR & CA & ASR & CA & ASR & CA \\
        \midrule
        BadNets-white & 1.30 & 88.27 & 1.22 & 94.05 & 0.96 & 98.19 & 0.81 & 97.14 & 1.16 & 97.85 \\
        ISSBA & 0.04 & 91.78 & 0.00 & 96.78 & 0.33 & 98.35 & 0.16 & 98.15 & 0.00 & 98.10 \\
        \bottomrule
    \end{tabular}}
\end{table}

\textbf{Defence performance for varying \poisonrate} is shown in \Cref{tab:diff-poi-defended}. We test a wide range of poison rates to determine the effectiveness of \MyMethod under different attack settings. Overall, our \MyMethod is able to defend against backdoor attacks with both high and low \poisonrate. The decrease in CA when \poisonrate is low is due to the worse performance of $\poisonmoduleSymbolShort$ at detecting potentially poisoned samples as shown in \Cref{tab:poison_rate_variable}. We believe that better isolation methods for collating $\UnlearnSetSymbol$ \citep{vit-def-doan} will result in higher CA. 

\textbf{Effects of different \confidencethresh values} are shown in \Cref{tab:variable-conf-thresh}. As one of the important hyperparameters in our \MyMethod, varying $\confidencethreshSymbol$ does not significantly affect model performance for both CIFAR10 and TinyImageNet. Looking at the ``Poison'' and ``Clean'' columns of \Cref{tab:variable-conf-thresh}, we generally see less data (ether poisoned or clean) having maximum class probability above the \confidencethresh as $\confidencethreshSymbol$ increases. Based on this observation, we argue that the performance of \MyMethod remains stable even as $\UnlearnSetSymbol$ decreases in size. 
\begin{table}[!h]
    \caption{Performance of \MyMethod with $\confidencethreshSymbol\in \{0.9, 0.95, 0.99\}$. The values in the ``Poison'' and ``Clean'' columns correspond to the percentage of poisoned and clean data, respectively, that's classified as poisoned data by $\poisonmoduleSymbolShort$ for the corresponding $\confidencethreshSymbol$. Note that $\confidencethreshSymbol=0.95$ is the default setting. All values given in percentages. }
    \label{tab:variable-conf-thresh}
    \centering
    \vspace{-9pt}
    \scalebox{0.7}{
    \begin{tabular}{llcccccccccccc}
\toprule
 \multirow[c]{2}{*}{Dataset} & \multirow[c]{2}{*}{\large $\confidencethreshSymbol$}& \multicolumn{4}{c}{BATT} & \multicolumn{4}{c}{BadNets-white} & \multicolumn{4}{c}{Smooth} \\
 \cmidrule(lr){3-6} \cmidrule(lr){7-10} \cmidrule(lr){11-14} 
 &  & ASR & CA & Poison & Clean & ASR & CA & Poison & Clean & ASR & CA & Poison & Clean \\
\midrule
\multirow[c]{3}{*}{CIFAR10} & 0.90& 0.06 & 94.42 & 94.51 & 13.61 & 0.81 & 97.47 & 95.71 & 11.53 & 0.05 & 96.79 & 97.23 & 15.16 \\
 & 0.95& 0.02 & 98.23 & 95.83 & 7.10 & 0.96 & 98.19 & 95.00 & 5.62 & 0.09 & 97.77 & 95.81 & 7.79 \\
 & 0.99& 1.74 & 98.09 & 90.68 & 0.78 & 1.27 & 98.09 & 92.70 & 0.55 & 0.32 & 97.98 & 90.76 & 1.26 \\
\cline{1-14}
\multirow[c]{3}{*}{TinyImageNet} & 0.90& 6.73 & 65.31 & 92.68 & 0.52 & 0.12 & 65.10 & 89.45 & 0.48 & 0.00 & 66.03 & 92.27 & 0.88 \\
 & 0.95& 3.07 & 64.15 & 88.48 & 0.11 & 0.12 & 66.35 & 87.44 & 0.21 & 0.17 & 66.03 & 89.30 & 0.41 \\
 & 0.99& 1.10 & 65.41 & 92.19 & 0.00 & 0.12 & 65.35 & 82.03 & 0.01 & 2.45 & 64.65 & 80.17 & 0.07 \\
\cline{1-14}
\bottomrule
\end{tabular}}
\end{table}

\textbf{The complexity of the \poisonmodule} as represented by the depth of $\poisonmoduleSymbolShort$ significantly affects the defence performance of \MyMethod as shown in \Cref{tab:poison-module-depth}. As the depth of $\poisonmoduleSymbolShort$ increases, it becomes more complex and more adept at learning non-poisoned samples. Since $\poisonmoduleSymbolShort$ is confident about a larger number of clean images, this causes the number of clean data in $\UnlearnSetSymbol$ to become higher and reduces the amount of data learned by $\robustmoduleSymbolShort$. Therefore, as $\poisonmoduleSymbolShort$ becomes deeper, CA decreases since the \robustmodule unlearns more clean data (rows for CIFAR10 and TinyImageNet of \Cref{tab:poison-module-depth}). This effect is especially pronounced for simpler datasets (e.g., CIFAR10) because simpler datasets are more easily learned given the same model complexity, leading to more clean images being directed to $\UnlearnSetSymbol$. Tuning $\confidencethreshSymbol$ for different depth leads to better performance as shown in the last two rows of \Cref{tab:poison-module-depth}.
\begin{table}[!h]
    \centering
    \caption{Performance of \MyMethod with varying \poisonmodule depth. The $\confidencethreshSymbol$ values used for the last three rows are chosen after inspecting the distribution of maximum class probability values $\max(\sigma(\poisonmoduleSymbol{\Image})$ using poisoned training data. All values given in percentages.}
    \vspace{-9pt}
    \label{tab:poison-module-depth}
    \scalebox{0.9}{\begin{tabular}{lcccccccc}
    \toprule
     & \multirow[c]{2}{*}{Depth ($\confidencethreshSymbol$)} & \multicolumn{2}{c}{BadNets-white} & \multicolumn{2}{c}{Blended} & \multicolumn{2}{c}{ISSBA} \\
    \cmidrule(lr){3-4} \cmidrule(lr){5-6} \cmidrule(lr){7-8}
     & & ASR & CA & ASR & CA & ASR & CA \\
    \midrule
    \multirow[c]{3}{*}{CIFAR10} & 1 (0.95) & 0.96 & 98.19 & 0.00 & 98.27 & 0.33 & 98.35 \\
     & 2 (0.95) & 1.30 & 56.09 & 0.00 & 86.49 & 0.00 & 87.90 \\
     & 3 (0.95) & 0.00 & 18.04 & 0.00 & 64.83 & 0.00 & 34.21 \\
    \cline{1-8}
    \multirow[c]{3}{*}{TinyImagenet} & 1 (0.95) & 0.12 & 66.35 & 0.00 & 66.41 & 0.05 & 40.60 \\
     & 2 (0.95) & 0.02 & 60.26 & 0.00 & 63.77 & 0.04 & 37.68 \\
     & 3 (0.95) & 0.01 & 59.11 & 0.00 & 61.80 & 0.12 & 36.50 \\
    \cline{1-8}
    %\multirow[c]{3}{*}{\shortstack[l]{CIFAR10 \\ (Variable $\confidencethreshSymbol$)}} & 1 (0.95) & 0.96 & 98.19 & 0.00 & 98.27 & 0.33 & 98.35 \\
     \multirow[c]{2}{*}{\shortstack[l]{CIFAR10 \\ (Variable $\confidencethreshSymbol$)}} & 2 (0.99) & 0.92 & 97.65 & 0.00 & 92.10 & 0.00 & 98.25 \\
     & 3 (0.998) & 0.82 & 92.79 & 0.00 & 98.20 & 0.00 & 98.02 \\
    \cline{1-8}
    \bottomrule
    \end{tabular}}
\end{table}
\vspace{-0.5cm}
\subsection{Ablation Study on Defence Design}
\textbf{We demonstrate that \MyMethod works well for Vision Transformer variants and CNN architectures} in \Cref{tab:diff-vit-arch}. We evaluate \MyMethod where the following Vision Transformer variants are used as the \robustmodule: CaiT-XXS \citep{vit-variant-cait}, DeiT-S \citep{vit-variant-deit}, PiT-XS \citep{vit-variant-pit}, ViT-S (default architecture, \citet{vit-seminal-disovitskiy}), and XCiT-Tiny \citep{vit-variant-xcit}. In addition, we use ResNet-18 \citep{resnet-seminal} and WideResNet-50-2 \citep{wide-resnet-seminal} to evaluate our defence on non-ViT architectures. The Interleaved Ensemble Unlearning framework generally performs well for most architectures. In addition, \MyMethod trains high-performing models when $\alpha=0$ where $\FinetuneSetSymbol$ is clean (see ``No Attack'' column of \Cref{tab:diff-vit-arch}).

\textbf{The effects of using constant unlearning rate} is shown in \Cref{tab:const-unlearn-rate}. Our defence is slightly more effective when $\UnlearnLRSymbol$ and $\TuneLRSymbol$ differ by a small constant factor (first three rows of CIFAR10 \& TinyImageNet in \Cref{tab:const-unlearn-rate}). However, on average there is only a small difference between the performance of Dynamic and constant $\UnlearnLRSymbol$. For example, Dynamic $\UnlearnLRSymbol$ on average achieves $65.97\%$ CA on TinyImageNet, only $2.48$pp lower than the best CA at $\UnlearnLRSymbol=\TuneLRSymbol$; ASR is comparable. To have fewer hyperparameters, we use Dynamic $\UnlearnLRSymbol$ instead of $\UnlearnLRSymbol=c\cdot\TuneLRSymbol$ for hyperparameter $c$.

\begin{table}[!h]
    \centering
    \caption{Performance of \MyMethod with different model architectures using CIFAR10. The penultimate column showcases CA when applying \MyMethod on clean $\FinetuneSetSymbol$. The ``No Defence'' column uses BadNets-white as the attack and is tuned without defence. The first layer of ViT-S is used as the \poisonmodule for all models. We use $\confidencethreshSymbol=0.99$ for ``No Attack'' since this choice mounts an effective defence as shown in \Cref{tab:variable-conf-thresh}. All values given in percentages. }
    \label{tab:diff-vit-arch}
    \vspace{-9pt}
    \scalebox{0.83}{
    \begin{tabular}{lcccccccc@{\extracolsep{0.6cm}}c@{\extracolsep{0.0cm}}ccc}
    \toprule
     \multirow[c]{2}{*}{Variant} & \multicolumn{2}{c}{BATT} & \multicolumn{2}{c}{BadNets-white} & \multicolumn{2}{c}{ISSBA} & \multicolumn{2}{c}{Smooth} & \multicolumn{2}{c}{No Attack} & \multicolumn{2}{c}{No Defence} \\
     \cmidrule(lr){2-3} \cmidrule(lr){4-5} \cmidrule(lr){6-7} \cmidrule(lr){8-9} \cmidrule(lr){10-11} \cmidrule(lr){12-13}
      & ASR & CA & ASR & CA & ASR & CA & ASR & CA & \hspace{0.4cm}ASR\hspace{0.4cm} & CA & ASR & CA \\
    \midrule
CaiT-XXS & 0.01 & 97.01 & 0.98 & 97.07 & 0.76 & 97.11 & 1.20 & 97.09 & - & 97.40 & 96.80 & 97.09 \\
DeiT-S & 0.25 & 97.83 & 1.00 & 97.94 & 0.32 & 98.22 & 2.05 & 97.83 & - & 98.21 & 96.96 & 97.96 \\
PiT-XS & 0.09 & 96.44 & 1.14 & 96.62 & 4.73 & 96.74 & 3.53 & 96.66 & - & 96.78 & 96.70 & 96.66 \\
ViT-S & 0.02 & 98.14 & 0.93 & 97.95 & 0.06 & 98.15 & 0.09 & 97.61 & - & 97.87 & 97.34 & 98.12 \\
XCiT-Tiny & 0.96 & 87.72 & 0.99 & 87.67 & 1.04 & 85.47 & 4.66 & 83.96 & - & 92.48 & 95.07 & 91.79 \\
\cmidrule(lr){1-13}
ResNet-18 & 0.35 & 90.99 & 7.22 & 91.68 & 6.37 & 91.00 & 6.38 & 91.45 & - & 92.45 & 94.25 & 91.73 \\
VGG-11 & 3.19 & 88.86 & 8.39 & 90.22 & 11.63 & 89.45 & 9.70 & 73.78 & - & 91.22 & 95.36 & 90.59 \\
\cline{1-13}
\bottomrule
    \end{tabular}
    }
\end{table}
\begin{table}[!h]
    \centering
    \caption{Performance of \MyMethod using different ways of computing $\UnlearnLRSymbol$. Average ASR/CA across rows are shown in the last column. Dynamic $\UnlearnLRSymbol$ is computed using \Eqref{eqn:unlearn-rate}. All values given in percentages. }
    \vspace{-9pt}
    \label{tab:const-unlearn-rate}
    \scalebox{0.72}{
    \begin{tabular}{llcccccccccc@{\extracolsep{0.6cm}}c@{\extracolsep{0.0cm}}c}
\toprule
 \multirow[c]{2}{*}{Dataset} & \multirow[c]{2}{*}{$\UnlearnLRSymbol$} & \multicolumn{2}{c}{BATT} & \multicolumn{2}{c}{BadNets-white} & \multicolumn{2}{c}{Smooth} & \multicolumn{2}{c}{Trojan-WM} & \multicolumn{2}{c}{l0-inv} & \multicolumn{2}{c}{Average} \\
 \cmidrule(lr){3-4} \cmidrule(lr){5-6} \cmidrule(lr){7-8} \cmidrule(lr){9-10}\cmidrule(lr){11-12}\cmidrule(lr){13-14}
 &  & ASR & CA & ASR & CA & ASR & CA & ASR & CA & ASR & CA & \hspace{0.35cm}ASR\hspace{0.35cm} & CA \\
\midrule
\multirow[c]{4}{*}{CIFAR10} & $1\cdot \TuneLRSymbol$ & 0.02 & 98.02 & 0.72 & 97.97 & 0.05 & 98.01 & 0.00 & 93.84 & 0.00 & 97.76 & 0.20 & 96.96 \\
 & $2\cdot \TuneLRSymbol$ & 0.01 & 93.95 & 0.72 & 93.16 & 0.04 & 93.56 & 0.00 & 85.59 & 0.00 & 88.57 & 0.19 & 91.56 \\
 & $4\cdot \TuneLRSymbol$ & 0.00 & 79.37 & 0.63 & 83.16 & 0.01 & 70.43 & 0.00 & 70.50 & 0.00 & 76.45 & 0.16 & 75.87 \\
 & Dynamic & 0.02 & 98.23 & 0.96 & 98.19 & 0.09 & 97.77 & 0.00 & 98.15 & 0.00 & 98.19 & 0.27 & 98.09 \\
\cline{1-14}
\multirow[c]{4}{*}{TinyImageNet} & $1\cdot \TuneLRSymbol$ & 2.47 & 67.21 & 0.15 & 68.72 & 0.03 & 68.79 & 0.00 & 69.07 & 0.00 & 68.04 & 0.66 & 68.45 \\
 & $2\cdot \TuneLRSymbol$ & 0.02 & 66.87 & 0.05 & 63.68 & 0.02 & 63.66 & 0.00 & 67.04 & 0.00 & 60.93 & 0.02 & 65.31 \\
 & $4\cdot \TuneLRSymbol$ & 0.00 & 63.38 & 0.07 & 47.53 & 0.00 & 51.10 & 0.00 & 49.08 & 0.00 & 29.72 & 0.02 & 52.77 \\
 & Dynamic & 3.07 & 64.15 & 0.12 & 66.35 & 0.17 & 66.03 & 0.00 & 67.33 & 0.00 & 65.89 & 0.84 & 65.97 \\
\cline{1-14}
\bottomrule
\end{tabular}}
\end{table}

\section{Discussion and Limitations}
\label{sec:discussion}
The \textbf{major finding} is that the Interleaved Unlearning Framework is a high-performing defence for tuning benign models on backdoored datasets. The \textbf{impact} is that this framework is an \textbf{improvement for ViTs} in terms of stability and performance over existing unlearning-based methods that aim to cleanse models \textit{after} tuning on backdoored data. 
\begin{figure}[!h]
    \centering
    \includegraphics[width=0.9\linewidth]{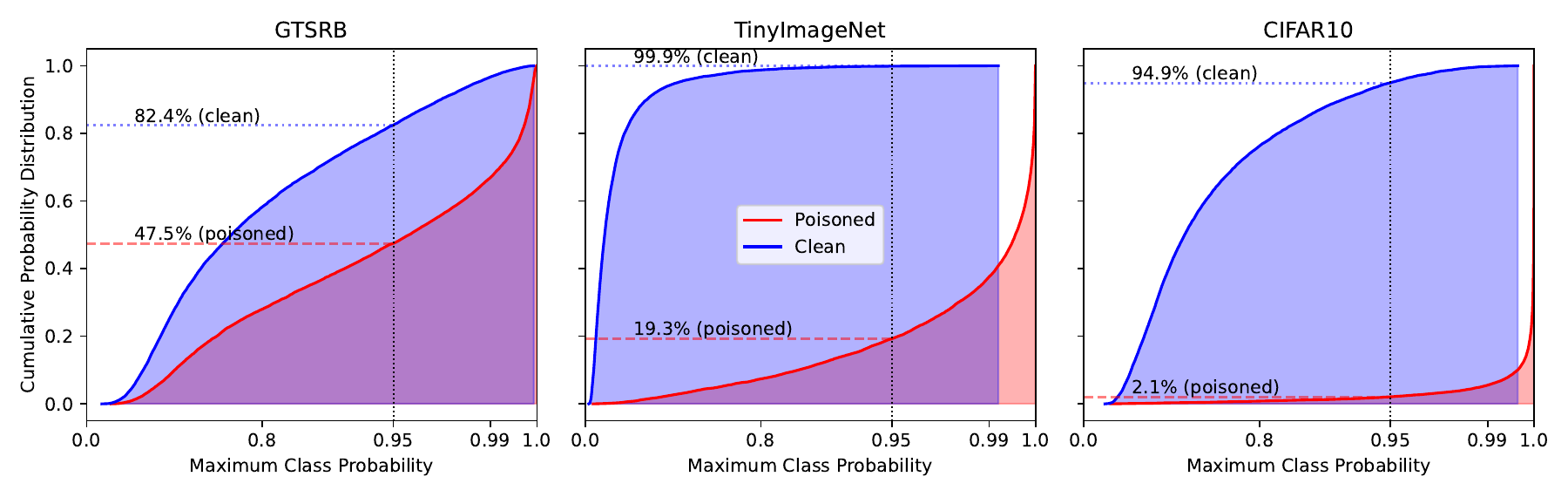}
    \vspace{-12pt}
    \caption{Maximum class probability $\max(\sigma(\poisonmoduleSymbol{\Image})$ CDF based on logits produced by the \poisonmodule on clean and poisoned data for the ISSBA attack on the three datasets. Dotted horizontal lines show percentages of clean/poisoned data whose $\max(\sigma(\poisonmoduleSymbol{\Image})$ lie below 0.95.}
    \label{fig:issba-cdf}
\end{figure}

Weakness [a] \textbf{Worse performance on less complex datasets} (e.g., GTSRB). Our \MyMethod fails against ISSBA when using the GTSRB dataset as shown in \Cref{tab:all-results} (cell coloured red) and underperforms on GTSRB in general. We suggest that this happens because the GTSRB dataset is easily learnt by $\poisonmoduleSymbolShort$. Evidence is shown in \Cref{fig:issba-cdf}, which plots the CDF of the maximum class probability values for poisoned/clean data for all three datasets using the ISSBA attack. Compared to CIFAR10 and TinyImageNet, clean images from the GTSRB dataset is only marginally more difficult to learn than backdoored GTSRB images. In \Cref{fig:issba-cdf}, a higher proportion of clean data has $\max(\sigma(\poisonmoduleSymbol{\Image})>0.95$ and a lower proportion of poisoned data has $\max(\sigma(\poisonmoduleSymbol{\Image})>0.95$ for GTSRB when compared to the other two datasets. Therefore, using a shallow ViT as the \poisonmodule is insufficient for discerning poisoned data from clean data for GTSRB.

Weakness [b] \textbf{The weaker the attack, the worse the defender's performance.} We believe that defending/detecting weak attacks is as important as defending strong attacks. A key difficulty in designing performant unlearning-based backdoor defence methods is identifying and mitigating weak attacks. An example of a relatively weak attack is WaNet \citep{backdoor-att-wanet}, and as shown in \Cref{tab:wanet-cifar10}, the two unlearning-based methods (our \MyMethod and ABL) we consider are less performant. Weakness [a] and [b] have a similar root cause. Both weaknesses are caused by a less effective $\poisonmoduleSymbolShort$ for isolating backdoored data. This effect is also seen to a smaller extent against the clean label SIG attack \citep{backdoor-att-sig-clean-label} on TinyImageNet, where the ASR without defence is $\approx68\%$. As shown in \Cref{tab:all-results}, although ASR for SIG when defended using \MyMethod is the lowest when comparing across different defences, the ASR for SIG is higher compared to the ASR on other attacks when defending using our \MyMethod.

\textbf{Solutions} for weaknesses [a] and [b]. A better isolation method can be used in place of $\poisonmoduleSymbolShort$, such as in \citet{vit-def-doan}. We surmise that using \citet{vit-def-doan}'s isolation method would make interleaved unlearning even more effective: although our \MyMethod does not require poisoned data isolation rate to be close to 100\% (see the ASR and CA values in \Cref{tab:variable-conf-thresh} where adjacent ``Poison'' values are $\approx 0.8$ and \Cref{tab:unlearning_set_size_variable} where $\UnlearnSetSizeSymbol_i\div\alpha\in \{0.5,0.9\}$), a more effective isolation method causes the \robustmodule to learn less backdoored data and unlearn less clean data. 

\begin{minipage}[t]{0.7\textwidth}
    \centering
    \vspace{-2.7cm}
        \captionof{table}{Performance of different backdoor defences on the WaNet attack \citep{backdoor-att-wanet} using the CIFAR10 dataset. Best and second-best values are bolded and underlined, respectively. All values are reported in percentages.}
        \label{tab:wanet-cifar10}
        \vspace{-9pt}
        \scalebox{0.9}{\begin{tabular}{cccccccc}
        \toprule
     \multicolumn{2}{c}{No Defence} & \multicolumn{2}{c}{I-BAU} & \multicolumn{2}{c}{ABL} & \multicolumn{2}{c}{\MyMethod} \\
     \cmidrule(lr){1-2}\cmidrule(lr){3-4}\cmidrule(lr){5-6}\cmidrule(lr){7-8}
     ASR & CA & ASR & CA & ASR & CA & ASR & CA \\
    \midrule
     72.81 & 97.39 & \textbf{10.26} & \underline{90.24} & 0.00 & 10.00 & \underline{26.01} & \textbf{94.17} \\
    \bottomrule
        \end{tabular}}
\end{minipage}%
\begin{minipage}[t]{0.3\textwidth}
    \centering
    \includegraphics[width=\textwidth,trim={0cm 0cm 0cm 0.9cm},clip]{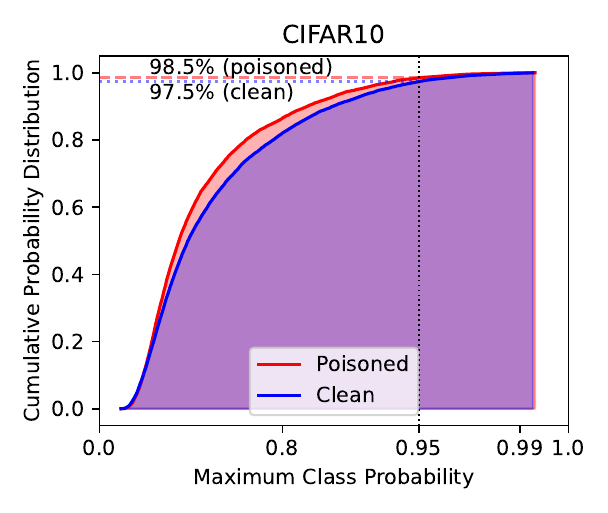}
    \vspace{-14pt}
    \captionof{figure}{$\max(\sigma(\poisonmoduleSymbol{\Image})$ CDF on clean/poisoned data for WaNet using CIFAR10.}
    \label{fig:wanet-cifar10}
\end{minipage}

Weakness [c] \textbf{Instability during defence} in stage 2 occurs when defending the VGG-11 model architecture against the Smooth attack. The instability causes NaN loss values during finetuning.

Potential \textbf{solutions} for weakness [c]: either (1) using $\UnlearnLRSymbol=c\cdot\TuneLRSymbol$ for hyperparameter $c$ instead of the Dynamic $\UnlearnLRSymbol$ explained in \Eqref{eqn:unlearn-rate} or (2) replacing $\CELoss{\robustmoduleSymbol{\potentialPoisonImage_{k-1}}}{\groundTruthOneHot_{k-1}}$ with a weighted moving average of successive cross entropy losses in \Eqref{eqn:unlearn-rate} may prevent instabilities from being introduced when alternating between finetuning and unlearning steps.

\section{Conclusion}
\label{sec:conclusion}
This work presents a novel and effective method for finetuning benign ViTs on backdoored datasets called Interleaved Ensemble Unlearning (\MyMethod). We use a small and shallow ViT (the \poisonmodule) to distinguish between clean and backdoored images and show that alternating between learning clean data and unlearning poisoned data during defence is an effective way of preserving high clean accuracy whilst foiling the backdoor attack. We demonstrate that our defence is effective for complicated real-world datasets and discuss ways to make \MyMethod more robust. 

\textbf{Impact}. This paper's impact goes beyond defending Vision Transformers against backdoor attacks. We believe that the Interleaved Ensemble Unlearning framework, which extends ABL \citep{abl-seminal} and Denoised PoE \citep{denoised-poe}, can be used to tune benign models with a great variety of different model architectures. In addition, we encourage future work to consider and remedy the weaknesses we point out in \Cref{sec:discussion} for unlearning-based backdoor defences.

\newpage

\bibliography{references}
\bibliographystyle{iclr2025_conference}

\newpage 
\appendix
\section{More Implementation Details}
\label{sec:more-implementation-details}
\subsection{Backdoor Attack Details}
\label{sec:appendix-attack-details}
\begin{figure}[!ht]
    \centering
    \includegraphics[width=0.95\linewidth]{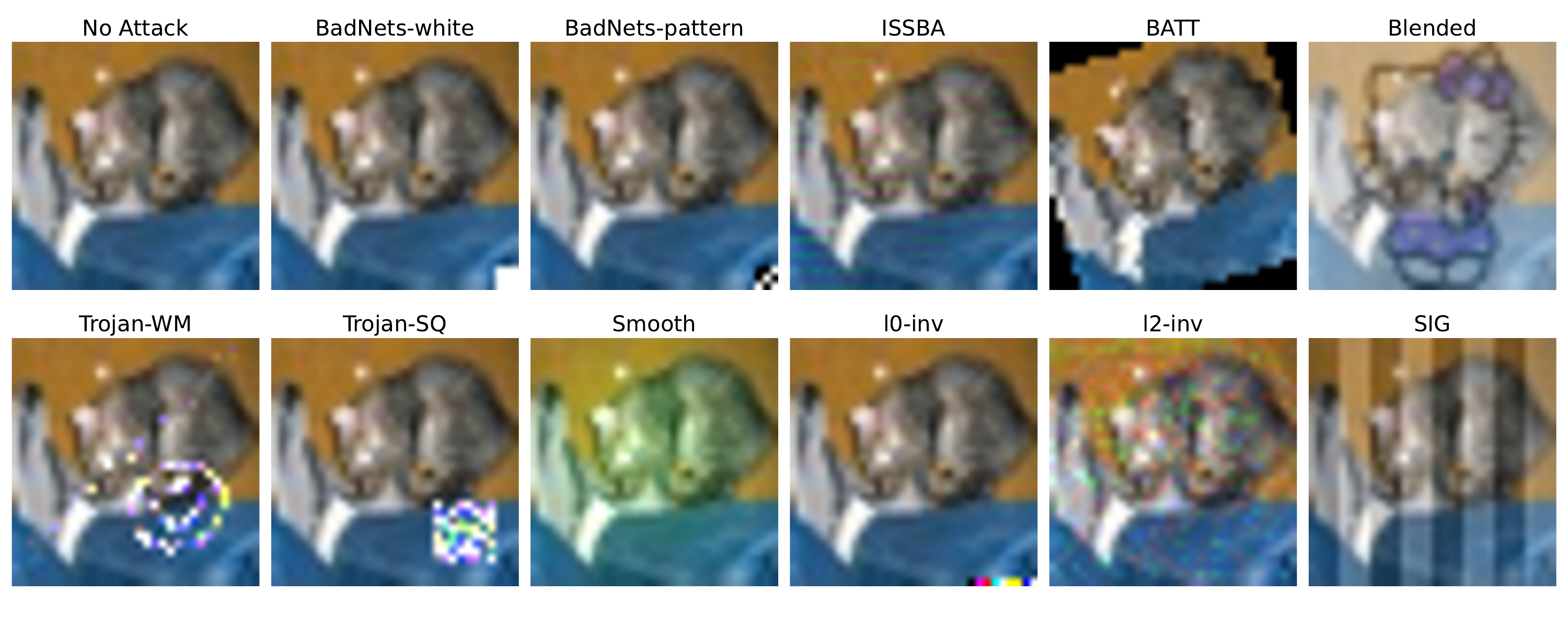}
    \vspace{-12pt}
    \caption{Visualisation of backdoored images (CIFAR10). }
    \label{fig:backdoor-images-vis}
\end{figure}

For SIG  \citep{backdoor-att-sig-clean-label} we poison 100\% of the chosen target class regardless of the dataset. We mostly base our data poisoning code on BackdoorBox's attacks \citep{backdoorbox-toolbox}; for attacks that are not available in BackdoorBox, we adapt our code from the attack authors' source code. We modify the code for BATT \citep{backdoor-att-batt} implemented in BackdoorBox by moving the attacker-specified transformation to before the data augmentation step. For ISSBA \citep{backdoor-att-issba}, one encoder/decoder pair is trained for each of the three datasets. \Cref{fig:backdoor-images-vis} shows visualisations of backdoored images on CIFAR10. Throughout the paper, the target class used by the attacker is class 1. 

\subsection{Backdoor Defence Details}
\label{sec:appendix-defence-details}
The data augmentations used for finetuning without defence and in our method (\MyMethod) are based on the augmentations in \citet{SiT} where \texttt{drop\_perc} and \texttt{drop\_replace} are set to 0.3 and 0.0, respectively. Three views of each image are produced: an unaugmented image, a clean crop with only colour jitter, and a corrupted crop with colour jitter and patch-based corruption. We used the pretrained checkpoint in \citet{SiT} for finetuning. Note that only the non-corrupted view is used to finetune models when demonstrating the effectiveness of \MyMethod with different model architectures in \Cref{tab:diff-vit-arch}. Each image's spatial dimension is $224\times224$ pixels. For all experiments except for those found in \Cref{tab:diff-vit-arch}, ViT-S is used as the base architecture: patch size of the ViT is $16\times16$; we use $\texttt{embed\_dim}=384, \texttt{ num\_heads}=6, \texttt{ mlp\_ratio}=4$, and LayerNorm \citep{layernorm-seminal}. 

\textbf{Finetuning without defence.} On all datasets, the ViT is finetuned for 10 epochs using the Adam \citep{adam-optimiser-seminal} optimizer with initial learning rate $2\cdot 10^{-5}$ and a cosine annealing learning rate scheduler that terminates at $1\cdot 10^{-6}$, weight decay using a cosine annealing scheduler starting from $0.04$ and increasing to $0.1$, and \textit{effective} batch size 128 (recall that the two augmented views of each image is used during finetuning). 
\begin{table}[!h]
    \centering
    \caption{Defence parameters used for CIFAR10/TinyImageNet and GTSRB. Parameters for CIFAR10/TinyImageNet have never been tuned (i.e., they were set to these values prior to running \textit{any} experiments with \MyMethod).}
    \label{tab:defence-param}
    \vspace{-9pt}
    \scalebox{0.9}{\begin{tabular}{lcccc}
        \toprule
        Dataset & Pre-finetuning Epochs/Warmup & Pre-finetuning lr & $\confidencethreshSymbol$ & lr decay\\
        \midrule 
        GTSRB & 5/No & $1\cdot 10^{-3}$ & $0.998$ & No\\
        CIFAR10/TinyImageNet & 10/Yes & $2\cdot 10^{-4}$ & $0.95$ & No\\
        \bottomrule
    \end{tabular}}
    
\end{table}

\textbf{Our method (\MyMethod). }For \textit{stage 1}, the differences in the hyperparameters used for CIFAR10/TinyImageNet and GTSRB to pre-finetune the \poisonmodule are shown in \Cref{tab:defence-param}. A weight decay of $0.0$ is applied to the \poisonmodule throughout stage 1 and no data augmentations are applied. If learning rate warmup is used, the learning rate scheduler performs a linear warmup for one epoch starting from $lr=0$. The batch size is 64. Recall that \poisonmodule is a shallow ViT of depth 1. For \textit{stage 2}, we use the same parameters during finetuning as for finetuning without defence. Please refer to \Cref{sec:method} and \Eqref{eqn:unlearn-rate} for an explanation of the unlearning rate $\UnlearnLRSymbol$.

\textbf{ABL.} Since the model architecture we use is different compared to the architectures used in \citet{abl-seminal}, we perform hyperparameter tuning using the BadNets-white attack on all three datasets. The \isolationratio $\isolationratioSymbol$ (fraction of $\FinetuneSetSymbol$ to unlearn) is set to 0.01 and every image in $\UnlearnSetSymbol$ is poisoned by default since we verify that LGA/Flooding can accurately select poisoned images on BadNets-white using CIFAR10. \Cref{tab:abl-hparam} shows the ABL hyperparameter tuning results for all three datasets. We choose $5\cdot 10^{-7}, 1\cdot 10^{-6}, 2\cdot 10^{-7}$ as the unlearning rate for CIFAR10, GTSRB, and TinyImageNet, respectively. In addition, we use the Adam optimiser.

\begin{table}[!h]
    \centering
       \caption{Hyperparameter tuning for ABL on CIFAR10, GTSRB, and TinyImageNet. Bolded are reasonably good values that correspond to our choices for unlearning lr.}
    \label{tab:abl-hparam}
    \vspace{-9pt}
    \scalebox{0.9}{\begin{tabular}{lcccccc}
\toprule
 \multirow[c]{2}{*}{Unlearning lr} & \multicolumn{2}{c}{CIFAR10} & \multicolumn{2}{c}{GTSRB} & \multicolumn{2}{c}{TinyImageNet} \\
 \cmidrule(lr){2-3} \cmidrule(lr){4-5} \cmidrule(lr){6-7}
 & ASR & CA & ASR & CA & ASR & CA \\
\midrule
$5.0 \cdot 10^{-8}$ & 97.29 & 98.40 & - & - & 97.80 & 61.34 \\
$1.0 \cdot 10^{-7}$ & 97.02 & 98.41 & - & - & 95.61 & 61.06 \\
$2.0 \cdot 10^{-7}$ & 96.19 & 98.38 & 94.53 & 95.56 & \textbf{0.25} & \textbf{59.29} \\
$3.0 \cdot 10^{-7}$ & 85.64 & 98.37 & 93.85 & 95.49 & 0.03 & 57.30 \\
$5.0 \cdot 10^{-7}$ & \textbf{9.63} & \textbf{98.08} & 83.84 & 95.46 & 0.00 & 45.52 \\
$1.0 \cdot 10^{-6}$ & 7.00 & 94.73 & \textbf{4.84} & \textbf{93.56} & 0.00 & 3.06 \\
$2.0 \cdot 10^{-6}$ & - & - & 0.00 & 83.11 & - & - \\
$3.0 \cdot 10^{-6}$ & - & - & 0.00 & 72.39 & - & - \\
$5.0 \cdot 10^{-6}$ & 0.00 & 84.78 & 0.00 & 36.74 & 0.00 & 0.50 \\
$1.0 \cdot 10^{-5}$ & 0.00 & 10.00 & 0.00 & 3.56 & 0.00 & 0.50 \\
$5.0 \cdot 10^{-5}$ & 0.00 & 10.00 & 0.00 & 3.56 & 0.00 & 0.50 \\
$1.0 \cdot 10^{-4}$ & 0.00 & 10.00 & 0.00 & 0.95 & 0.00 & 0.50 \\
\bottomrule
\end{tabular}}
    
\end{table}

\textbf{I-BAU.} We also perform hyperparameter tuning for I-BAU \citep{ibau} for the same reasons above. Following suggestions in the appendix of \citet{backdoor-def-t-and-r}, we tune \texttt{outer\_lr} $\in \{5\cdot10^{-4}, 1\cdot10^{-4}, 5\cdot10^{-5}, 1\cdot10^{-5}, 5\cdot10^{-6}\}$ and \texttt{inner\_lr} $\in \{0.1, 1, 5, 10, 20\}$ for CIFAR10 and TinyImageNet. GTSRB is not separately tuned since good eprformance is reached using CIFAR10's hyperparameters. We choose as the hyperparameters $(\texttt{outer\_lr}, \texttt{inner\_lr})=[(5\cdot 10^{-5}, 5), (5\cdot 10^{-5}, 5), (5\cdot 10^{-5}, 10)]$ for CIFAR10, GTSRB, and TinyImageNet, respectively. In addition, we use the Adam optimiser as the outer optimiser for I-BAU. For every dataset, $5000$ images from the \textit{testing set} are used for the unlearning step (\texttt{unlloader} in their code). Note that $5000$ clean images taken from the \textit{testing set} is the default setup in the defence code of I-BAU.

\textbf{AttnBlock.} This is the defence referred to as the ``\textit{Attn Blocking}'' defence in \citet{backdoor-def-attn-block}. We use GradRollout \citep{GradRollout} to compute the interpretation map on image $\Image$ $\mathbf{I}_{\text{map}}(\Image)$ using the backdoored checkpoint and find the coordinates of the interpretation map's maximum $\max(\mathbf{I}_{\text{map}}(\Image)$. A $30\times30$ patch centred at the coordinates $\max(\mathbf{I}_{\text{map}})$ is zeroed out from the original image $\Image$ to form $\Image'$. If this centred patch goes outside of the image, the patch is shifted so that it is on the image's border. Then, the backdoored checkpoint is used to classify $\Image'$. The results for the test-time interpretation-informed defence proposed in \citet{backdoor-def-attn-block} is shown in \Cref{tab:attn-block-results}. Generally, the defence does not defend against the non-patch-based attacks evaluated in our work. 

\begin{table}[!h]
    \centering
    \caption{Performance of AttnBlock}
    \label{tab:attn-block-results}
    \vspace{-9pt}
    \scalebox{0.9}{\begin{tabular}{lcccccc}
\toprule
  \multirow[c]{2}{*}{Attack} & \multicolumn{2}{c}{CIFAR10} & \multicolumn{2}{c}{GTSRB} & \multicolumn{2}{c}{TinyImageNet} \\
  \cmidrule(lr){2-3} \cmidrule(lr){4-5} \cmidrule(lr){6-7}
 & ASR & CA & ASR & CA & ASR & CA \\
\midrule
BadNets-white & 34.71 & 98.21 & 29.05 & 94.73 & 54.35 & 60.21 \\
BadNets-pattern & 12.61 & 98.06 & 22.03 & 94.2 & 7.55 & 61.2 \\
ISSBA & 100.0 & 97.93 & 100.0 & 94.13 & 99.27 & 61.82 \\
BATT & 99.99 & 98.0 & 100.0 & 94.63 & 99.99 & 65.42 \\
Blended & 99.99 & 98.15 & 100.0 & 94.8 & 100.0 & 68.94 \\
Trojan-WM & 99.98 & 98.13 & 99.98 & 91.35 & 99.97 & 68.82 \\
Trojan-SQ & 99.48 & 98.09 & 99.47 & 94.06 & 99.63 & 61.97 \\
Smooth & 99.72 & 97.99 & 99.71 & 94.66 & 99.17 & 67.54 \\
l0-inv & 100.0 & 98.1 & 100.0 & 95.14 & 100.0 & 62.1 \\
l2-inv & 99.99 & 98.2 & 100.0 & 92.77 & 99.78 & 64.33 \\
SIG & 98.35 & 88.53 & 99.43 & 90.42 & 68.36 & 70.41 \\
\cmidrule(lr){1-7}
\textbf{Average} & 85.89 & 97.22 & 86.33 & 93.72 & 84.37 & 64.8 \\
\bottomrule
\end{tabular}}
    
\end{table}

\subsection{Hardware Resources}
Most experiments are conducted on one NVIDIA A100 GPU. A few experiments are (and can be) conducted on one NVIDIA Quadro RTX 6000 GPU. We did not reproduce I-BAU \citep{ibau} on the RTX 6000 GPU due to GPU memory constraints. 

\newpage
\section{Algorithm for Stage 2}
\label{sec:algorithms}
The steps for Stage 2 of \MyMethod is shown in \Cref{alg:stage2}.
\begin{algorithm}
\caption{Stage 2 of \MyMethod: Defending $\robustmoduleSymbolShort$}\label{alg:stage2}
    \begin{algorithmic}[1]
    \State \textbf{Input}: tuned \poisonmodule $\poisonmoduleSymbol{\cdot}$, potentially poisoned finetuning set $\FinetuneSetSymbol$, pretrained benign $\robustmoduleSymbol{\cdot}$, optimizer for finetuning $\text{Optim}^{\text{tune}}$, optimizer for unlearning $\text{Optim}^{\text{ul}}$, number of epochs, and learning rate schedule $\TuneLRSymbol$.
    \State \textbf{Output:} tuned \robustmodule $\robustmoduleSymbol{\cdot}$ that is benign.
    \For{every epoch}
        \State Initialise $\UnlearnSetSymbol$ to be an empty queue to store potentially poisoned images.
        \For{every batch $(\Image_b, \groundTruthOneHot_b)\in \FinetuneSetSymbol$} \Comment{The subscript `b' indicates a batch.}
        \State Compute $\logit_b$ using \eqref{eqn:weight-masking}, $\poisonmoduleSymbolShort$, and $\robustmoduleSymbolShort$ for this batch.
        \State Compute the cross entropy loss $\CELoss{\logit_b}{\groundTruthOneHot_b}$.
        \State Update $\vtheta_r$ using $\text{Optim}^{\text{tune}}$ to optimise for \Eqref{eqn:stage2-optim}.
        \State Update $\TuneLRSymbol$ based on the learning rate schedule.
        \State Add all $(\Image_{b,i},\groundTruthOneHot_{b,i})\in(\Image_b, \groundTruthOneHot_b)$ that satisfies $\max(\sigma(\poisonmoduleSymbol{\Image_{b,i}}))>\confidencethreshSymbol$ to $\UnlearnSetSymbol$.
        \If{a batch $(\potentialPoisonImage_b, \groundTruthOneHotPoison_b)\in \UnlearnSetSymbol$ is ready}
            \State Compute $\CELoss{\robustmoduleSymbol{\potentialPoisonImage_b}}{\groundTruthOneHotPoison_b}$ and $\UnlearnLRSymbol$ based on \Eqref{eqn:unlearn-rate}.
            \State Update $\vtheta_r$ using $\text{Optim}^{\text{ul}}$ to optimise for \Eqref{eqn:stage2-optim-unlearn} and record $\UnlearnLRSymbol$ for the next batch.
            \State Dequeue the current batch $(\potentialPoisonImage_b, \groundTruthOneHotPoison_b)$ from $\UnlearnSetSymbol$.
        \EndIf
        \EndFor
    \EndFor
    \end{algorithmic}
\end{algorithm}

\newpage
\section{Ablation Study (cont'd)}
\label{sec:appendix-ablation}

\textbf{Noisy logits}. \Cref{tab:noisy-logits} shows the effects of adding normally distributed zero-mean noise to $\poisonModuleLogit$ when accumulating the unlearn set. Whether noise is added or not has almost no effect on the model performance. 

\begin{table}[!h]
    \centering
    \caption{Performance of \MyMethod on CIFAR10 when adding normally distributed zero-mean noise with different variances to $\poisonModuleLogit$ \textit{after} computing $\weightMask$, i.e., $\weightMask$ is still defined according to \Eqref{eqn:weight-masking} but $\max(\sigma(\poisonModuleLogit+\mathbf{n})) > \confidencethreshSymbol$ where $\mathbf{n}\sim\gN(0,\sigma^2)$ is used to determine whether $\Image$ belongs in $\UnlearnSetSymbol$. This creates a mismatch between the data added onto $\UnlearnSetSymbol$ (unlearned by $\robustmoduleSymbolShort$) and the data learned by $\robustmoduleSymbolShort$.}
    \label{tab:noisy-logits}
    \vspace{-9pt}
    \begin{tabular}{lcccccccc}
\toprule
\multirow[c]{2}{*}{Variance} & \multicolumn{2}{c}{BATT} & \multicolumn{2}{c}{BadNets-white} & \multicolumn{2}{c}{ISSBA} & \multicolumn{2}{c}{Smooth} \\
 & ASR & CA & ASR & CA & ASR & CA & ASR & CA \\
\midrule
0.0 & 0.02 & 98.23 & 0.96 & 98.19 & 0.33 & 98.35 & 0.09 & 97.77 \\
0.1 & 0.02 & 98.09 & 0.93 & 98.23 & 0.10 & 98.43 & 0.08 & 97.85 \\
0.5 & 0.02 & 98.21 & 0.88 & 98.13 & 0.35 & 98.36 & 0.04 & 97.64 \\
1.0 & 0.01 & 98.15 & 0.95 & 98.09 & 0.05 & 98.30 & 0.05 & 97.91 \\
2.0 & 0.02 & 96.82 & 0.94 & 98.17 & 0.46 & 98.30 & 0.06 & 97.95 \\
\bottomrule
\end{tabular}
\end{table}

\textbf{The effect of using LGA/Flooding in conjunction with our \poisonmodule} is shown in \Cref{tab:lga-flooding-no-method}. The short-hand $\poisonmoduleSymbolShort$ \& $M$ means applying method $M$ when tuning $\poisonmoduleSymbolShort$ during stage 1 of our method. Given the high FNR for most attacks with CIFAR10 and TinyImageNet when using LGA/Flooding together with our $\poisonmoduleSymbolShort$, we argue that using \poisonmodule is orthogonal to LGA/Flooding for isolating poisoned data.
\begin{table}[!h]
    \centering
    \caption{Comparison of the three methods' (LGA, Flooding, no-method) ability to distinguish between poisoned and clean data when used in conjunction with our $\poisonmoduleSymbolShort$ during prefinetuning. The $\poisonmoduleSymbolShort$ is prefinetuned for 10 epochs using default parameters. The flooding/LGA parameter is set to $\gamma=1.5,\gamma=3.0$ for CIFRA10 and TinyImageNet, respectively. Each cell shows the FPR/FNR values as percentages (positive means ``poisoned''). }
    \vspace{-9pt}
    \label{tab:lga-flooding-no-method}
    \scalebox{0.9}{
    \begin{tabular}{lcccccc}
        \toprule
         \multirow[c]{2}{*}{Method} & \multicolumn{3}{c}{CIFAR10} & \multicolumn{3}{c}{TinyImageNet} \\
         \cmidrule(lr){2-4} \cmidrule(lr){5-7}
          & BadNets-white & ISSBA & Smooth & BadNets-white & ISSBA & Smooth \\
        \midrule
        $\poisonmoduleSymbolShort$ \& Flooding & 0.00/99.98 & 0.00/57.76 & 0.00/99.54 & 0.03/20.07 & 0.01/64.66 & 0.12/15.89 \\
        $\poisonmoduleSymbolShort$ \& LGA & 0.00/99.98 & 0.00/65.84 & 0.00/97.86 & 0.03/20.07 & 0.12/67.32 & 0.12/15.89 \\
        $\poisonmoduleSymbolShort$ \& no-method & 5.62/5.30 & 5.58/2.10 & 7.90/3.14 & 0.21/11.97 & 0.15/22.59 & 0.42/9.83 \\
        \bottomrule
    \end{tabular}
    }
\end{table}
\end{document}